\documentclass[lettersize,journal]{IEEEtran}
\usepackage{amsmath,amsfonts}
\usepackage{algorithmic}
\usepackage{algorithm}
\usepackage{array}
\usepackage[caption=false,font=normalsize,labelfont=sf,textfont=sf]{subfig}
\usepackage{textcomp}
\usepackage{stfloats}
\usepackage{url}
\usepackage{verbatim}
\usepackage{graphicx}
\usepackage{multirow}
\usepackage{makecell}
\usepackage{cite}
\usepackage{bm}
\usepackage{color}
\usepackage{booktabs}
\usepackage[colorlinks,
linkcolor=black,
anchorcolor=black,
citecolor=black]{hyperref}
\usepackage{balance}

\newcounter{daggerfootnote}

\begin{document}
\title{MASH: Cooperative-Heterogeneous Multi-Agent Reinforcement Learning for Single Humanoid Robot Locomotion}

\author{
    Qi Liu$^1$$^\dagger$, Xiaopeng Zhang$^2$$^\dagger$, Mingshan Tan$^2$, Shuaikang Ma$^2$, Jinliang Ding$^1$, Yanjie Li$^2$$^*$
\thanks{
    This work was supported by Shenzhen Basic Research Program (Grant No. JCYJ20220818102415033,  KJZD2023092311422045). \textit{(Corresponding author: Yanjie Li, autolyj@hit.edu.cn)}
}
\thanks{
    1 Faculty of Robot Science and Engineering, Northeastern University, Shenyang, 110819, China.
}
\thanks{
    2 Guangdong Key Laboratory of Intelligent Morphing Mechanisms and Adaptive Robotics and School of Intelligence Science and Engineering, the Harbin Institute of Technology Shenzhen, Shenzhen, 518055, China.
}
\thanks{$^\dagger$ These authors contributed equally to this work.}
}



\maketitle

\begin{abstract}
This paper proposes a novel method to enhance locomotion for a single humanoid robot through cooperative-heterogeneous multi-agent deep reinforcement learning (MARL). While most existing methods typically employ single-agent reinforcement learning algorithms for a single humanoid robot or MARL algorithms for multi-robot system tasks, we propose a distinct paradigm: applying cooperative-heterogeneous MARL to optimize locomotion for a single humanoid robot. The proposed method, \textbf{m}ulti-\textbf{a}gent reinforcement learning for \textbf{s}ingle \textbf{h}umanoid locomotion (MASH), treats each limb (legs and arms) as an independent agent that explores the robot’s action space while sharing a global critic for cooperative learning. Experiments demonstrate that MASH accelerates training convergence and improves whole-body cooperation ability, outperforming conventional single-agent reinforcement learning methods. This work advances the integration of MARL into single-humanoid-robot control, offering new insights into efficient locomotion strategies.
\end{abstract}


\section{Introduction}
\label{Section: Introduction}
\IEEEPARstart{D}{eep} reinforcement learning (RL) has achieved remarkable success in various domains of robotic control \cite{he2024hover,10508809}, such as quadruped locomotion \cite{margolis2022walk,10161144}, bipedal walking \cite{benbrahim1997biped,10611621}, and autonomous aerial vehicle navigation \cite{8004441}. This work focuses on advancing deep RL for humanoid robot locomotion, a complex yet critical challenge in robotics.

The locomotion methodologies for humanoid robots can be categorized into two principal paradigms: model-based locomotion methods and model-free learning-based methods. The former, exemplified by whole-body dynamics locomotion methods incorporating model predictive control and trajectory optimization \cite{10286076,sleiman2023versatile,da2019combining}, demonstrates robust performance in structured environments but relies on precise dynamic modeling. The latter encompasses deep reinforcement learning (RL) \cite{radosavovic2024real,kim2024not} and imitation learning \cite{chi2023diffusion}. Recent advances \cite{ha2024learning,jiang2024learning,haarnoja2024learning} have revealed the superior policy generalization capabilities of learning-based methods over model-based locomotion methods. Learning-based locomotion methods can be summarized into two classes: (1) phased policy training and system integration for the upper and lower body, and (2) whole-body motion via imitation learning or single-agent deep RL.

Phased policy training and system integration methods employ a modular decomposition and progressive integration framework, mitigating locomotion complexity in high-degree-of-freedom (DoF) systems. Despite significant advancements, this class method faces persistent challenges: Phased training may lead to insufficient coordination between upper-body and lower-body policies, limiting whole-body synergy. And its adaptability remains constrained for complex tasks and scenarios \cite{jiang2024learning,liu2024visual}. 

Whole-body imitation learning and single-agent deep RL methods  \cite{darvish2023teleoperation} begin by collecting or generating whole-body motion data for humanoid robots, then use the acquired data to train locomotion policies through imitation learning or single-agent deep RL. However, obtaining high-quality humanoid motion data faces significant challenges, including high collection costs, low efficiency, and complex post-processing requirements. Although imitation learning-based methods are straightforward, the resulting policies exhibit limited generalization capabilities. Compared to imitation learning, deep RL-based methods show improved generalization but fail to leverage deep RL's trial-and-error learning mechanism fully. Because these methods rely on human motion data as reference trajectories, they struggle to explore action and skill spaces beyond the demonstrated data, ultimately constraining their performance ceiling in complex environments. 

When applying deep RL to single-humanoid robot locomotion, the predominant method employs single-agent deep RL algorithms \cite{10610286,10341908}. However, these methods may exhibit limitations in addressing coordination challenges inherent in complex robotic systems. Existing solutions typically utilize either single-agent RL for individual robots or multi-agent deep reinforcement learning (MARL) for cooperative multi-robot tasks \cite{chen2022towards}. However, there remains an unexplored potential in leveraging MARL principles for enhanced coordination within a single humanoid robotic entity. Thus, developing more efficient whole-body cooperative locomotion methods to enhance humanoid robots’ mobility-manipulation coordination in unstructured environments remains challenging.

Cooperative MARL algorithms have demonstrated remarkable success in multi-agent cooperation in game artificial intelligence domains \cite{rashid2020monotonic,10466624}. Liu et al. \cite{liu2024masq} propose a novel method to improve locomotion learning for a single quadruped robot using MARL. Departing from conventional robot locomotion learning approaches, this paper proposes a novel method that models locomotion learning as a cooperative-heterogeneous MARL problem and uses the cooperative-heterogeneous MARL algorithm to enhance a single humanoid robot's locomotion. Our method enables superior coordination in complex tasks by treating each limb (two arms and two legs) as an independent agent within a cooperative MARL framework. The proposed cooperative-heterogeneous \textbf{m}ulti-\textbf{a}gent reinforcement learning for \textbf{s}ingle \textbf{h}umanoid locomotion (MASH) leverages a shared learning structure, where agents (limbs) collectively optimize locomotion through experience sharing and cooperative policy learning. 

The main contributions of this paper are as follows:
\begin{itemize}
    \item We propose MASH, a novel framework that reformulates humanoid locomotion as a cooperative-heterogeneous MARL problem. MASH enables more efficient coordination learning than conventional single-agent RL by treating each limb (arms and legs) as an independent agent with distinct action spaces.
    \item Experimental results show that the proposed method achieves superior gait execution and final performance, improves training efficiency and sample complexity, and enhances robustness in dynamic environments. These results validate the effectiveness of applying MARL principles to single humanoid robot control.
\end{itemize}

The remainder of this paper is structured as follows. Section \ref{Section: Related work} reviews related works. Section \ref{Section: Background} introduces the preliminaries of the Markov decision process (MDP) and RL. Section \ref{Section: Multi-Agent Reinforcement Learning for Single Humanoid Robot Locomotion} presents the proposed MASH method in detail. Section \ref{Section: Experiments} reports the experimental results that demonstrate the effectiveness of the proposed approach. Finally, Section \ref{Section: Conclusion and Future Work} concludes the paper and outlines directions for future work.

\section{Related work}
\label{Section: Related work}

\subsection{Deep RL for Humanoid Robot Locomotion}
\label{Subsection: Deep RL for Humanoid Robot Locomotion}
The locomotion methodologies for humanoid robots can be categorized into two principal paradigms: model-based locomotion methods and model-free learning-based methods. Model-based locomotion methods, exemplified by whole-body dynamics locomotion methods incorporating model predictive control and trajectory optimization \cite{10286076,sleiman2023versatile,da2019combining}, demonstrate robust performance in structured environments but rely on precise dynamic modeling. Model-free learning-based methods encompass deep reinforcement learning (RL) \cite{radosavovic2024real,kim2024not} and imitation learning \cite{chi2023diffusion}. Recent advances \cite{ha2024learning,jiang2024learning,haarnoja2024learning} have shown the superior policy generalization capabilities of learning-based methods over model-based methods. Learning-based locomotion methods can be summarized into two classes: (1) phased policy training and system integration for the upper and lower body, and (2) whole-body locomotion learning based on imitation learning or single-agent deep RL.

\textbf{Phased policy training and system integration methods} employ a modular decomposition and progressive integration framework, mitigating locomotion complexity in high degrees of freedom (DoFs) systems. This class method consists of three stages: (1) Decoupled policy training: The whole-body control problem is partitioned into independent policy training for the lower body (focused on locomotion stability) and the upper body (emphasizing manipulation dexterity) \cite{cheng2024expressive}. (2) System integration: Upper-body manipulation tasks are gradually incorporated after the lower-body policy converges \cite{lu2025mobile}. (3) Task-oriented optimization: Task constraints (e.g., end-effector trajectory tracking) are superimposed onto foundational mobility, with whole-body control achieved through coupled dynamic optimization \cite{he2024hover,zhangwococo}. Despite significant advancements, this class method faces persistent challenges: Decoupled policy training may lead to insufficient coordination between upper-body and lower-body policies, limiting whole-body synergy, and its adaptability remains constrained for complex tasks and scenarios \cite{jiang2024learning,liu2024visual}. Thus, developing more efficient whole-body cooperative locomotion methods to enhance humanoid robots’ mobility-manipulation coordination remains challenging.

\textbf{Whole-body locomotion learning based on imitation learning or single-agent deep RL methods} 
begin by collecting or generating whole-body motion data for humanoid robots \cite{darvish2023teleoperation}, then use the acquired data to train locomotion policies through imitation learning or single-agent deep RL. Based on data sources, this class method can be categorized into four types: (1) Robot teleoperation data collection \cite{darvish2023teleoperation,he2024learning}: Involves direct operation of physical robots to acquire kinematically accurate motion data. Although this yields high-quality, platform-specific data, it suffers from high acquisition costs and hardware dependence. (2) Human motion capture data: Methods like HumanPlus \cite{fu2024humanplus} record human motions using mocap systems and map them to the robot joint space. Although intuitive, this approach requires specialized equipment, incurs scalability challenges, and must address human-robot kinematic discrepancies. (3) Internet-sourced human video data: OKAMI \cite{li2024okami} extracts human motions from online videos, performs 3D reconstruction, and transfers them to robots. Although data is abundant, this method struggles with noisy inputs and kinematic mismatches. (4) Synthetic animation data: Works like OmniH2O \cite{he2024omnih2o} and PMCP \cite{luo2023perpetual} generate motions through animation tools and remap them to robots. Although flexible, such data may lack physical realism.

Furthermore, obtaining high-quality humanoid motion data faces significant challenges, including high collection costs, low efficiency, and complex post-processing requirements. Although imitation learning-based control methods are relatively straightforward, the resulting policies exhibit limited generalization capabilities. Compared to imitation learning, deep RL-based methods show improved generalization but fail to leverage deep RL's trial-and-error learning mechanism fully. Because these methods rely on human motion data as reference trajectories, they struggle to explore action and skill spaces beyond the demonstrated data, ultimately constraining their performance ceiling in complex environments.

\textbf{Other methods:} \cite{li2023robust} employs traditional trajectory optimization to generate reference trajectories for robots, combined with single-agent deep RL for trajectory tracking control. Although this approach has shown success in quadruped robotics, its application to humanoid robots remains constrained due to their higher DoF and more complex dynamic constraints. Some methods employ extensive reward shaping to guide policy learning, including hand-tuned motion tracking \cite{xie2018feedback}, periodic reward composition \cite{siekmann2021sim}, adversarial motion priors \cite{peng2021amp}, and periodic sinusoidal trajectories \cite{gu2024advancing}. However, designing reward functions manually is often labor-intensive and time-consuming, requiring extensive domain expertise and iterative tuning to achieve desired behaviors.

\subsection{MARL for Multi-robot Control}
\label{Subsection: MARL for multi-robot control}
MARL has demonstrated remarkable success across diverse multi-robot systems, including cooperation robot swarms \cite{blais2023reinforcement}, autonomous driving \cite{zhou2021smarts}, unmanned aerial vehicles coordination \cite{lim2022optimal}, and intelligent warehouse \cite{liu2024multi}. These applications showcase MARL's capacity to coordinate complex behaviors in physically embodied systems, where decentralized decision-making must reconcile environmental constraints with inter-agent coordination. In contrast to these multi-robot applications, this paper presents a novel paradigm by formulating single-humanoid locomotion as a cooperative MARL problem. Our method treats each leg and arm as an independent agent, departing from conventional methods that either model the robot as a single unified agent or focus on multi-robot cooperation.

\section{Background}
\label{Section: Background}
This section summarizes the MDP \cite{sutton2018reinforcement} and RL. The MDP considered in this paper is modeled as a tuple $(\mathcal{S}, \mathcal{A}, \mathcal{P}, \mathcal{R}, \gamma, T)$, where $\mathcal{S}$ is the state space, $\mathcal{A}$ is the action space, $\mathcal{P}: \mathcal{S} \times \mathcal{A} \times \mathcal{S} \rightarrow [0,1]$ represents the state transition probability. $\mathcal{R}: \mathcal{S} \times \mathcal{A} \rightarrow R$ is the reward function, $\gamma \in [0,1)$ is the discount factor, and $T$ denotes the time horizon. At each timestep, $t$, an action $a_{t} \in \mathcal{A}$ is selected according to a policy. The agent then transitions to the next state by sampling from $p\left(s_{t+1} \mid s_{t}, a_{t}\right)$, where $p \in \mathcal{P}$, and receives a scalar reward $r\left(s_{t}, a_{t}\right) \in \mathcal{R}$. The agent continues interacting with the environment until it reaches a terminal state. The goal of RL is to learn a policy $\pi: \mathcal{S} \times \mathcal{A} \rightarrow [0, 1] $ that maximizes the expected discounted cumulative rewards. For any policy $\pi$, the state-action value function ($Q$ function) is defined as follows:
\begin{equation}
    Q^{\pi}(s, a)={\mathbb{E}^{\pi}}\left[\sum_{t=0}^{T} \gamma^{t} r\left(s_{t}, a_{t}\right) \mid s_{0}=s, a_{0}=a\right]
    \label{Eq: Q function}
\end{equation}

Proximal Policy Optimization (PPO) \cite{schulman2017proximal} addresses key stability challenges in policy gradient methods through a constrained optimization approach. The algorithm's core innovation lies in its clipped surrogate objective function:
\begin{equation}
    L^{\textit{CLIP}}(\theta) = \mathbb{E}_t \left[ \min \left( r_t(\theta) \hat{A}_t, \text{clip}(r_t(\theta), 1 - \epsilon, 1 + \epsilon) \hat{A}_t \right) \right]
\end{equation}
where
\begin{equation}
    r_t(\theta) = \frac{\pi_{\theta}(a_t | s_t)}{\pi_{\theta_{\text{old}}}(a_t | s_t)}
\end{equation}
and
\begin{equation}
    \hat{A}_t = Q^{\pi}(s_t, a_t) - V_{\psi}(s_t)
\end{equation}
where $\psi$ denotes the parameters of value function ($V_{\psi}$) network, $\epsilon$ denotes a coefficient. The policy parameters $\theta$ are updated as follows:
\begin{equation}
    \theta \leftarrow \theta + \alpha \nabla_{\theta} L^{\textit{CLIP}}(\theta)
\end{equation}
PPO's constrained updates stabilize training and improve performance, making it practical for complex single-agent RL tasks.

\section{Multi-Agent Reinforcement Learning for Single Humanoid Robot Locomotion}
\label{Section: Multi-Agent Reinforcement Learning for Single Humanoid Robot Locomotion}
This paper proposes a novel MARL framework to enhance single-robot locomotion by leveraging inter-limb coordination. Our approach treats each limb (arms and legs) of the humanoid robot as an independent agent within a cooperative-heterogeneous multi-agent system, where agents share a global critic while maintaining individual observations and policies. This framework bridges the gap between single humanoid robot locomotion learning and multi-agent coordination learning, demonstrating that cooperative multi-agent learning strategies can improve a single humanoid robot's locomotion.

\begin{figure}[htbp]
    \centerline{\includegraphics[width=9cm,height=4.5cm]{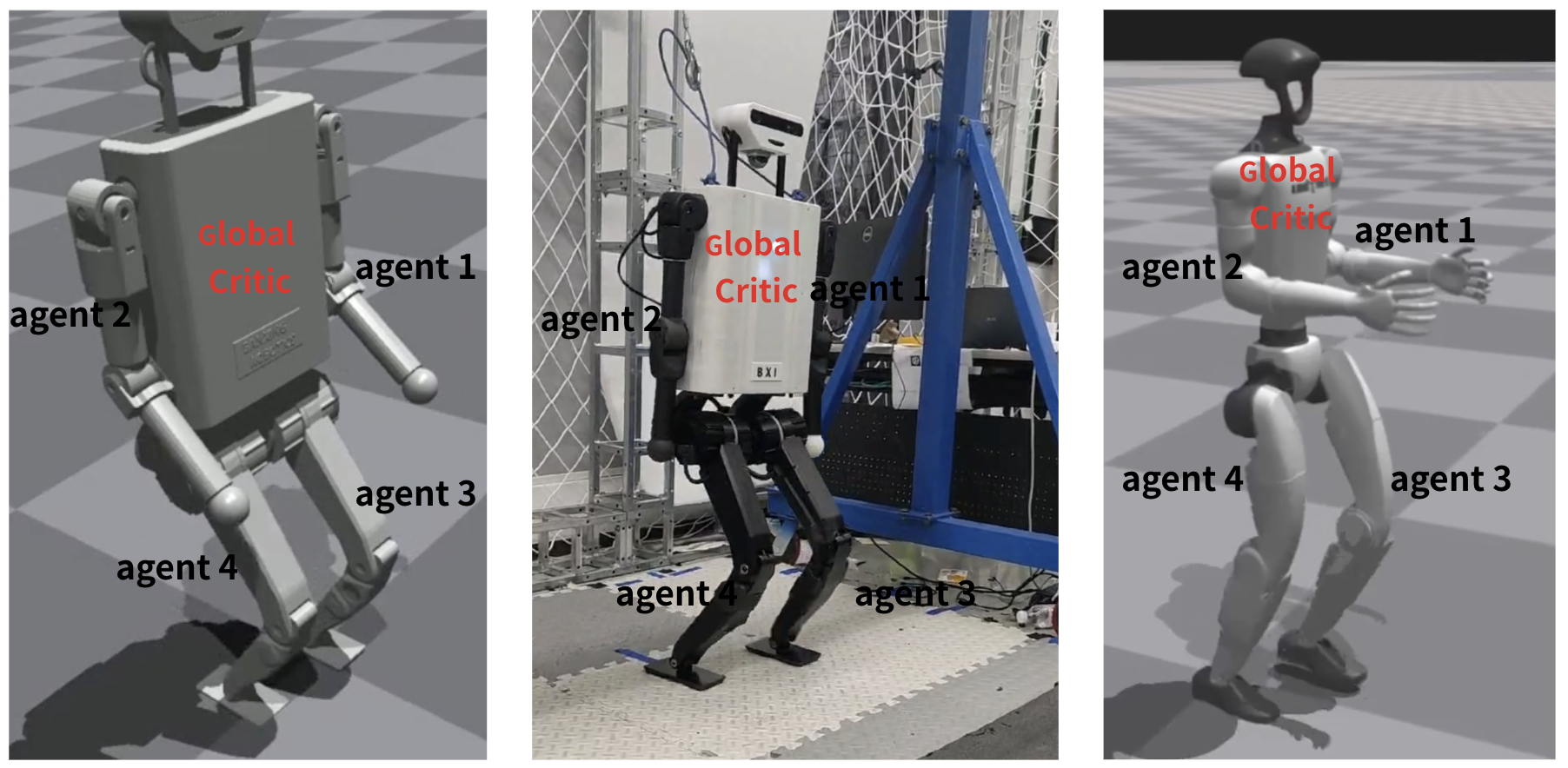}}
    \caption{MARL model for a single humanoid robot's locomotion}
    \label{Figure: MARL model for a single humanoid robot's locomotion}
\end{figure}

\begin{figure*}[htbp]
    \centerline{\includegraphics[width=17cm, height=11cm]{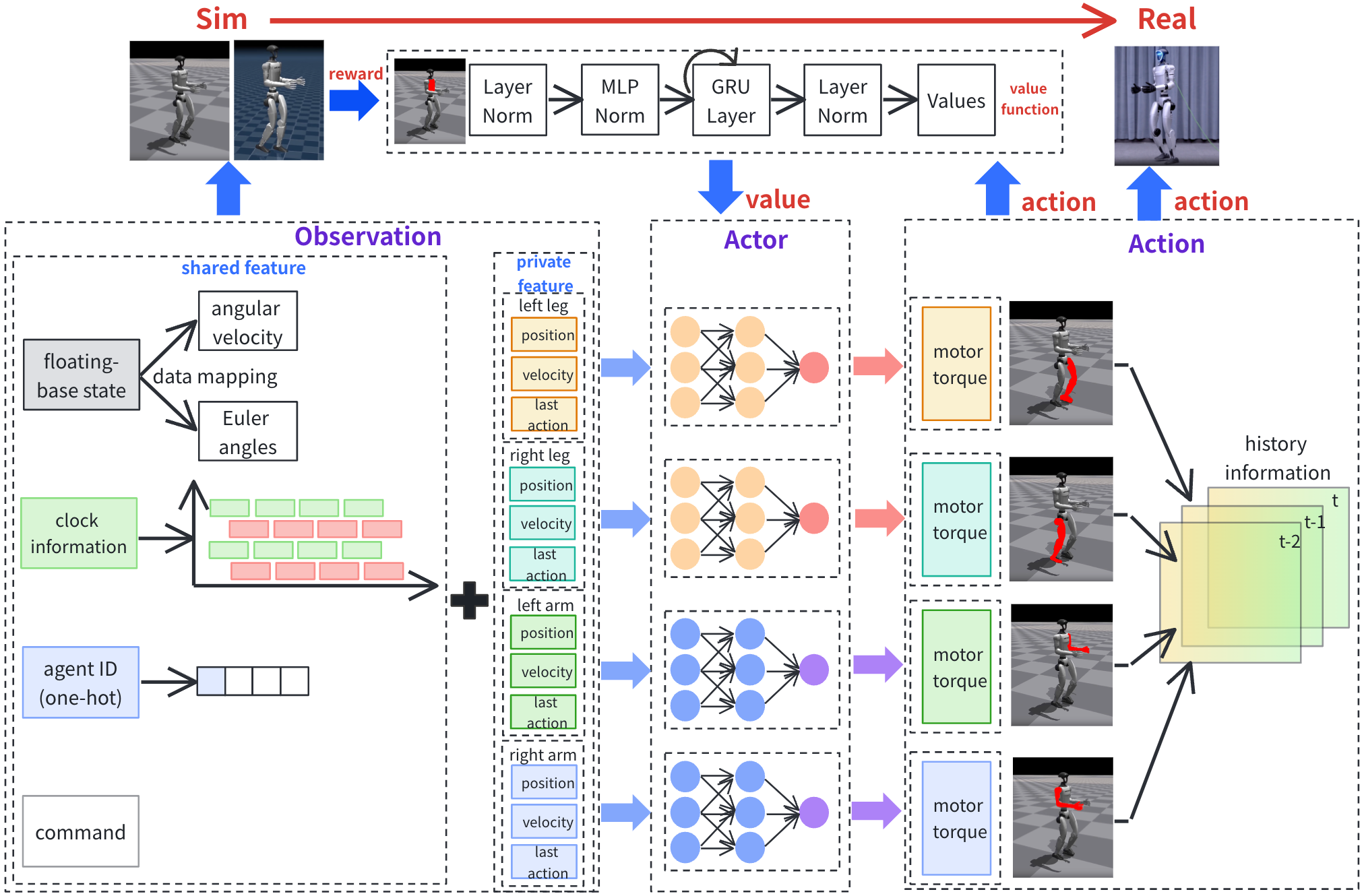}}
    \caption{The framework of MASH}
    \label{Figure: The framework of MASH}
\end{figure*}

\subsection{MASH}
\label{Subsection: MASH}
This paper models a single humanoid robot locomotion as a cooperative multi-agent problem, which is described as a partially observable decentralized Markov decision process (decPOMDP) \cite{ong2009pomdps}. The decPOMDP is defined by the tuple $G=(\mathcal{S}, \mathcal{A}, \mathcal{P}, r, \mathcal{Z}, \mathcal{O}, N, \gamma, T)$. $\mathcal{S}$ is the state space, $\mathcal{A}$ is the action space, $\mathcal{P}$ is the state transition distribution, $r$ is the reward function, $\mathcal{Z}$ is the observation space, $\mathcal{O}$ is the observation function, $N$ is the number of agents, $\gamma$ is the discount factor, and $T$ is the time horizon. At each time step $t$, each agent $n \in \{1, \ldots, N\}$ selects an action $a^n_t \in \mathcal{A}$, resulting in a joint action $\bm{a_t} = \{a^1_t, a^2_t, \ldots, a^N_t\}$. The environment transitions to a new state $s_{t+1}$ according to $\mathcal{P}(s_{t+1} | s_t, \bm{a_t})$ and provides a shared reward $r(s_t, \bm{a_t})$. Each agent receives an observation $z^n_t$ from $\mathcal{O}(s_t, n)$ and maintains an observation-action history $\tau^n_t$. MARL aims to learn policies $\{\pi^n\}_{n=1}^N$ that maximize expected cumulative rewards:
\begin{equation}
    J(\pi)=\mathbb{E}\left[\sum_{t=0}^{T} \gamma^{t} r(s_t, \bm{a_t})\right]
    \label{Eq: MARL objective function}
\end{equation}

This paper proposes MASH, which applies the MARL algorithm to treat different parts of a single humanoid robot as independent agents, trained collaboratively using shared rewards. Specifically, this paper uses multi-agent PPO \cite{yu2022surprising} (MAPPO) to solve the modeled multi-agent problem. MAPPO optimizes the following objective function in a multi-agent system:
\begin{equation}
\begin{aligned}
    L^{\text{\textit{CLIP}}}_{\text{\textit{MAPPO}}}(\theta) &= \sum_{i=1}^{n} \mathbb{E}_t \left[ \min \left( r_t^i(\theta_i) \hat{A}_t, \right. \right. \\
    & \left. \left. \text{clip}(r_t^i(\theta_i), 1 - \epsilon, 1 + \epsilon) \hat{A}_t \right) \right]
\end{aligned}
    \label{Eq: MAPPO loss}
\end{equation}
where 
\begin{equation}
    r_t^i(\theta_i) = \frac{\pi_{\theta_i}(a_t^i | o_t^i)}{\pi_{\theta_{i,\text{old}}}(a_t^i | o_t^i)}
    \label{Eq: MAPPO ratio}
\end{equation}
where $i$ denotes the $i$-th agent in MARL. Each agent updates its policy parameters as follows:
\begin{equation}
    \theta_i \leftarrow \theta_i + \alpha \nabla_{\theta_i} L^{\textit{CLIP}}_i(\theta_i)
    \label{Eq: MAPPO policy parameters update}
\end{equation}

This work adopts the centralized training with decentralized execution (CTDE) paradigm \cite{foerster2017stabilising} to address key challenges in MARL while ensuring training stability in dynamic environments. The implementation employs dual neural networks that separately learn a policy ($\pi_\theta$) and an enhanced value function $\left(V_\psi(s)\right)$, where the latter incorporates global state information to improve training stability. By leveraging this architecture, the proposed approach can achieve superior final performance, accelerated convergence rates, and enhanced deployment robustness compared to conventional methods, demonstrating particular efficacy for complex robotic control tasks requiring both coordinated learning and operational flexibility.

\textbf{State Space and Observation:} A shared-parameter actor network is constructed for both the bipedal and dual-arm systems, respectively, receiving concatenated observations from the two agents. For the bipedal system, each agent's observation includes: motor positions \( q_t \in \mathbb{R}^6 \) (representing the angles of the 6 leg motors), motor velocities \( \dot{q_t} \in \mathbb{R}^6 \) (representing the rotational speeds of the motors), the action from the previous time step \( a_{t} \in \mathbb{R}^6 \), trigonometric timing guidance \( \phi_t \in \mathbb{R}^2 \) for gait sequencing (which regulates the timing of steps), the Euler angles \( \theta_{t} \in \mathbb{R}^3 \) and angular velocities \( \omega_{t} \in \mathbb{R}^3 \) of the torso, control commands \( c_t \in \mathbb{R}^4 \) provided by the remote controller (including a binary standing label, velocities in the \( x \) and \( y \) directions, and the yaw angular velocity), a one-hot encoding \( e_t \in \mathbb{R}^2 \) for each agent. Thus, each agent has a 32-dimensional observation, resulting in a total input dimension of \( o_t^{\text{actor}} \in \mathbb{R}^{64} \) (32 × 2) for the actor network. For the dual-arm system, there are only 4 degrees of freedom for the motors. As a result, each arm agent receives a 26-dimensional observation, leading to a total input dimension of \( o_t^{\text{actor}} \in \mathbb{R}^{52} \) (26 × 2).

The Critic network takes the global observations of the entire robot body as input, including: motor positions \( q_t \in \mathbb{R}^{20} \), motor velocities \( \dot{q_t} \in \mathbb{R}^{20} \), the action from the previous time step \( a_{t} \in \mathbb{R}^{20} \), motor position deviation \( d_t \in \mathbb{R}^{20} \), timing guidance \( \phi_t \in \mathbb{R}^2 \) for the limbs, control commands \( c_t \in \mathbb{R}^4 \) provided by the remote controller, as well as the linear velocity \( v_{t} \in \mathbb{R}^3 \), Euler angles \( \theta_{t} \in \mathbb{R}^3 \) and angular velocities \( \omega_{t} \in \mathbb{R}^3 \) of the torso. In addition, the observation includes the external disturbance forces \( f_t \in \mathbb{R}^2 \) and torques \( \tau_t \in \mathbb{R}^3 \), environment friction coefficients \( \mu_t \in \mathbb{R}^1 \), body mass \( m_t \in \mathbb{R}^1 \), and two binary masks representing the stance state and contact state of the limbs, each \( \in \mathbb{R}^2 \). Therefore, the total input dimension for the Critic network is \( o_t^{\text{critic}} \in \mathbb{R}^{106} \).  

\textbf{Action Space:} The output of the bipedal actor network is a continuous action \( a_t \in \mathbb{R}^{12} \), representing the torques for the 12 motors. For the dual-arm system, the actor network outputs \( a_t \in \mathbb{R}^{8} \).  

The Critic network outputs the value function \( V_t \in \mathbb{R}^4 \), which is used to compute the advantage function.

\textbf{Reward Function:} The reward functions in Table \ref{tab: rewards} are designed to optimize the robot's performance by encouraging desired behaviors and penalizing undesired ones. Key rewards include: \textit{joint position} to follow reference postures; velocity tracking terms such as \textit{tracking linear velocity} and \textit{tracking angular velocity} to follow commanded motions; energy-related penalties including \textit{DOF torques}, \textit{DOF velocity}, and \textit{DOF acceleration} to encourage smooth and efficient movement; gait-specific rewards like \textit{feet air time}, \textit{feet clearance}, and \textit{feet contact number} promote coordinated stepping, while stability and safety are addressed by \textit{orientation}, \textit{collision}, \textit{feet slip}, and \textit{base height}. Additional terms such as \textit{action smoothness} and \textit{torque rate} help ensure motion consistency and reduce actuation spikes. Together, these terms shape the robot’s actions toward achieving robust and physically plausible behavior in complex tasks.

\begin{table}[htbp]
    \centering
    \caption{Reward function}
    \label{tab: rewards}
    \setlength{\tabcolsep}{4pt} 
    \renewcommand{\arraystretch}{1.8} 
    \resizebox{0.5\textwidth}{!}{ 
    \begin{tabular}{l|l|l}
        \hline
        \multicolumn{3}{c}{\uppercase{\textbf{Reward Settings, Corresponding Equations, and Their Scales}}} \\
        \hline
        \textbf{Reward Term} & \textbf{Equation} & \textbf{Scale} \\
        \hline
        Joint Position & $\exp\left(-\|\boldsymbol{q}-\boldsymbol{q}_\mathrm{default}-\boldsymbol{q}_\mathrm{ref}\|\right)$ & $3.5$ \\
        Tracking Linear Velocity & $\exp \left( -\frac{\|\mathbf{v}_{\text{cmd}} - \mathbf{v}_{b}\|^2}{\sigma_{t}} \right)$ & $1.5$\\
        Tracking Angular Velocity & $\exp \left( -\frac{\left(\mathbf{\omega}_{\text{cmd}, z} - \mathbf{\omega}_{b, z}\right)^2}{\sigma_{\text{yaw}}} \right)$ & $1.4$\\
        DOF Torques & $\sum\left(\frac{\tau}{\tau_{\max}}\right)^2$ & $-2.0 \times 10^{-3}$\\
        DOF Velocity & $\sum_i \|\mathbf{v}_{d, i}\|^2$ & $-5 \times 10^{-4}$\\
        DOF Acceleration & $\sum_i \left( \frac{\mathbf{v}_{d, i, t} - \mathbf{v}_{d, i, t-1}}{\Delta t} \right)^2$ & $-1.0 \times 10^{-7}$ \\
        Feet Air Time & $\sum_{i=1}^2t_{\mathrm{air},i}\cdot\exp\left(-\alpha\cdot\|\Delta\mathbf{x}_i\|\right)$ & $2.0$\\
        Feet Clearance & $\sum(|z_{feet}-h_{\mathrm{target\_feet}}|<\delta)$ & $2.0$\\
        Feet Contact Number & $\sum \begin{cases} +1, & \mathrm{if~contact}_i=\mathrm{stance}_i \\ -0.3, & \mathrm{otherwise} \end{cases}$ & $1.2$\\
        Orientation & $\exp(|\theta_\text{pitch, roll}|)+\exp(\|g_\mathrm{proj}\|)$ & $1.0$\\
        Collision & $\sum \left(1.0 \cdot (\|\mathbf{f}_{c}\| > 0.1)\right)$ & $-1.0$ \\
        Feet Slip & $\sum (c_f \cdot \|\mathbf{v}_{f}\|^2)$ & $-5 \times 10^{-2}$ \\
        Base Height & $\exp\left(-|h_{base}-h_{target_base}|\right)$ & $0.2$\\
        Action Smoothness 1 & $\sum \left( \mathbf{a}_t - \mathbf{a}_{t-1} \right)^2$ & $-0.1$ \\
        Action Smoothness 2 & $\sum \left( \mathbf{a}_t - 2\mathbf{a}_{t-1} + \mathbf{a}_{t-2} \right)^2$ & $-0.1$ \\
        Torque Rate & $\sum\left(\frac{\tau_t-\tau_{t-1}}{\tau_{\max}\cdot\Delta t}\right)^2$ & $-2 \times 10^{-4}$\\
        \hline
    \end{tabular}
    }
\end{table}

\subsection{Multi-agent Actor and Global Critic Networks}
\label{Subsection: Multi-agent actor networks}

In the Isaac Gym \cite{makoviychuk2021} simulation environment, the robot receives observations and rewards to facilitate learning. Observations are divided into shared and private features, with both actor and critic networks using rewards to optimize the policy. 

The actor-network employs an MLP/RNN architecture with an activation layer that generates actions and their log probabilities, while the critic network utilizes a similar MLP/RNN structure to estimate state values. During training, the actor-network selects actions based on environmental observations, with the critic providing value estimates to guide policy updates. 
Action outputs correspond to desired joint positions. A proportional-derivative (PD) controller computes the corresponding joint torques based on these targets, which are then applied to the robot model in the physics simulator.
Upon completion of training, the optimized actor-network is directly implemented on the robotic platform for real-world operation. The complete learning pipeline is depicted in Fig. \ref{Figure: The framework of MASH}.

We model each leg and arm as an independent agent for the humanoid robot system while employing a shared-parameter actor-network across two limbs and two arms, respectively (that is, two arms share one shared-parameter actor-network, while two legs share another). This architectural choice offers two key advantages: (1) it significantly reduces computational overhead compared to maintaining separate networks for each limb, and (2) it properly captures the inherent symmetry and coordinated nature between the left and right limbs of a humanoid robot during locomotion. Unlike conventional multi-agent systems like StarCraft \cite{rashid2020monotonic}, where agents operate independently, our humanoid's legs and arms form an intrinsically coupled system, symmetrically arranged around the body's center and mechanically constrained to move in coordination. The shared-parameter network naturally encodes these physical symmetries and coordination requirements, making it particularly well-suited for robotic control while maintaining the benefits of a MARL framework.

We express the policy for each leg and arm as follows:
\begin{equation}
    \pi_{\theta_i}^{leg}(a_{i,t} \mid s_{i,t}) = \pi_{\theta}^{leg}(a_{i,t} \mid s_{i,t})
    \label{Eq: Shared Parameter Network}
\end{equation}
\begin{equation}
    \pi_{\theta_i}^{arm}(a_{i,t} \mid s_{i,t}) = \pi_{\theta}^{arm}(a_{i,t} \mid s_{i,t})
    \label{Eq: Shared Parameter Network for Arms}
\end{equation}
where \text{$i$ = 1, 2} indexes the left and right limbs. The leg policy $\pi_{\theta_i}^{leg}$ is shared across both legs, and the arm policy $\pi_{\theta_i}^{arm}$ is shared across both arms.

To enhance the coordination among the agents, we augment each independent observation with shared observations, including Euler angle and angular velocity calculated from inertial measurement unit data, the temporal director, and the agent identifier (ID). The temporal director $T_i(t)$ guides the gait sequence of each leg and coordinates the motion patterns of each arm under different movement postures, while the agent ID is necessary for the shared-parameter network. This setup ensures the independence of each agent while improving their cooperative capabilities.\label{observation}
The temporal director helps to synchronize the movements of different legs and coordinate the motions of the arms, ensuring smooth and balanced gait patterns. It can be defined as follows:
\begin{equation}
    T_i(t) = \sin\left( 2\pi(k t + \Delta_i) \right)
    \label{Eq: temporal director}
\end{equation}
where
\begin{itemize}
    \item \( k \) is the scaling factor of gait cycle.
    \item \(\Delta_i\) is the phase offset for the \(i\)-th limb (leg or arm), which determines its relative timing within the gait cycle to ensure coordinated movement.
\end{itemize}

The global critic employs a centralized value function within the CTDE framework, utilizing global observations that combine individual limb observations and shared system-wide information. This architecture enables effective coordination among the two-legged and two-armed agents while maintaining the benefits of single-agent optimization approaches like PPO. The critic network processes this comprehensive state representation to learn value functions that guide policy updates. This ensures synchronized learning across all agents while accounting for the robot's global state and inter-limb dependencies.

\subsection{Sim-to-Real}
\label{Subsection: Sim-to-real}
To enhance sim-to-real transfer, we employ domain randomization. To enhance sim-to-real transfer, we employ domain randomization. Based on the timing of the application, all domain randomization parameters can be categorized into two types. The first type includes physical parameters randomized at environment initialization, such as link mass and inertia, joint damping and friction, ground friction, and gravity. The second type involves parameters randomized at each simulation step, including action delay, torque noise, and external perturbations. These randomizations enhance the policy's robustness and adaptability to varying physical conditions.

\section{Experiments}
\label{Section: Experiments}

We conducted MASH experiments using the BanXing humanoid robot with various experiments. Section \ref{Subsection: Experiment Setup} proposes the experimental setup. Section \ref{Subsection: Simulation Experiments} shows the experimental results of the simulation. Section \ref{Subsection: Real-world Experiments} shows the real-world experiments and comparisons.

\begin{figure}[htbp]
    \centerline{\includegraphics[width=9cm,height=4.5cm]{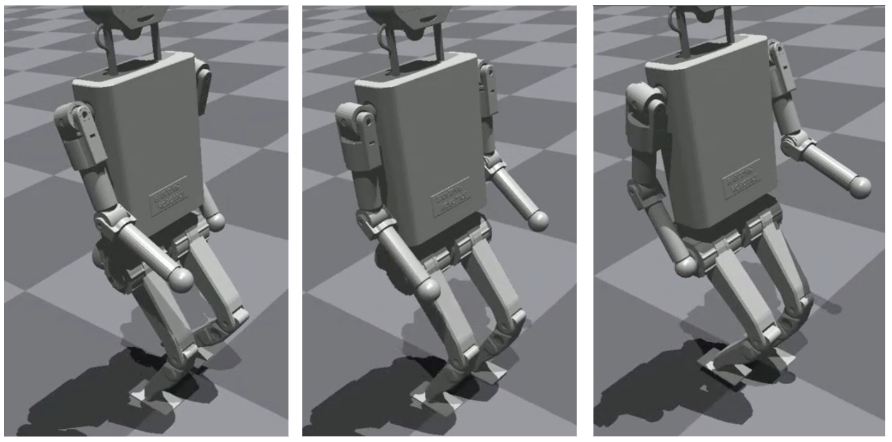}}
    \caption{Simulation experiments of MASH}
    \label{Figure: Simulation experiments of MASH}
\end{figure}

\subsection{Experiment Setup}
\label{Subsection: Experiment Setup}

\subsubsection{Task Description}
\label{Subsubsection: Task Description}
In the simulation environment, we train a single humanoid robot to walk forward on flat terrain in a fast and stable manner. This task serves as a standard benchmark in humanoid robot control, assessing the robot's capabilities in terms of speed and balance. Our training pipeline begins with upright locomotion training in Isaac Gym, which includes: (i) \textbf{bipedal walking} (with both legs forming a single group of multiple agents) and (ii) \textbf{arm-swing walking} (with both legs and both arms forming two groups of multiple agents). Compared to bipedal walking, arm-swing walking imposes higher requirements on limb coordination. We then perform Sim-to-Sim transfer to the MuJoCo simulator to rigorously assess the robustness of the learned policy under different simulation environments. Ultimately, the policy is deployed on a physical robot to demonstrate its effectiveness in real-world scenarios.

\subsubsection{Evaluation Metrics}
\label{Subsubsection: Evaluation Metrics}
To comprehensively demonstrate the superiority of MASH, we establish the following evaluation metrics:
\begin{itemize}
    \item \textbf{Convergence time} \(T_{Conv}\): we define the convergence time as the number of training iterations required for the average episodic reward to stably reach 95\% of the asymptotic reward value during training.
    \item \textbf{Action smoothness} \(S_{action}\): we measure the continuity and stability of control actions (e.g., joint torques or target joint positions), aiming to avoid abrupt changes in the action sequence. Quantified by the squared L2-norm of the second-order difference of the action sequence:
    \begin{equation}
    S_{action}=\frac{1}{T}\sum_{t=1}^{T-1}\sum_{i=1}^{N}(a_{i,t+1}-a_{i,t})^2
    \label{Eq: action smoothness}
    \end{equation}
    \item \textbf{Torso stability} \(S_{torso}\): The stability of the torso during locomotion is evaluated by measuring the fluctuations in torso height and orientation, including pitch, roll, and yaw angles: 
    \begin{equation}
    S_{torso}=w_h\cdot\mathrm{Var}(h_t)+w_\theta\cdot\mathrm{Var}(\theta_{t})
    \label{Eq: torso stability}
    \end{equation}
    \item \textbf{Limb coordination} \(C_{limb}\): By analyzing the motion trajectories of key joints in the legs, such as the hip or knee, and computing their relative phase difference over the gait cycle, limb coordination can be quantitatively evaluated: 
    \begin{equation}
    C_{limb}=\frac{1}{T}\sum_{t=1}^T|\phi_{\mathrm{left}}(t)-\phi_{\mathrm{right}}(t)-\phi_{\mathrm{target}}|
    \label{Eq: torso stability}
    \end{equation}
\end{itemize}

\subsubsection{Baselines}
\label{Baselines}
To better demonstrate the strong potential of MASH, we compare it with the conventional single-agent PPO algorithm, training both for 3000 episodes in the simulation environment with an episode length of 48 steps.

\begin{figure*}[htbp]
    \centering
    \subfloat[]{  
    \begin{minipage}[b]{0.5\textwidth} 
        \centering
        \includegraphics[width=0.9\linewidth]{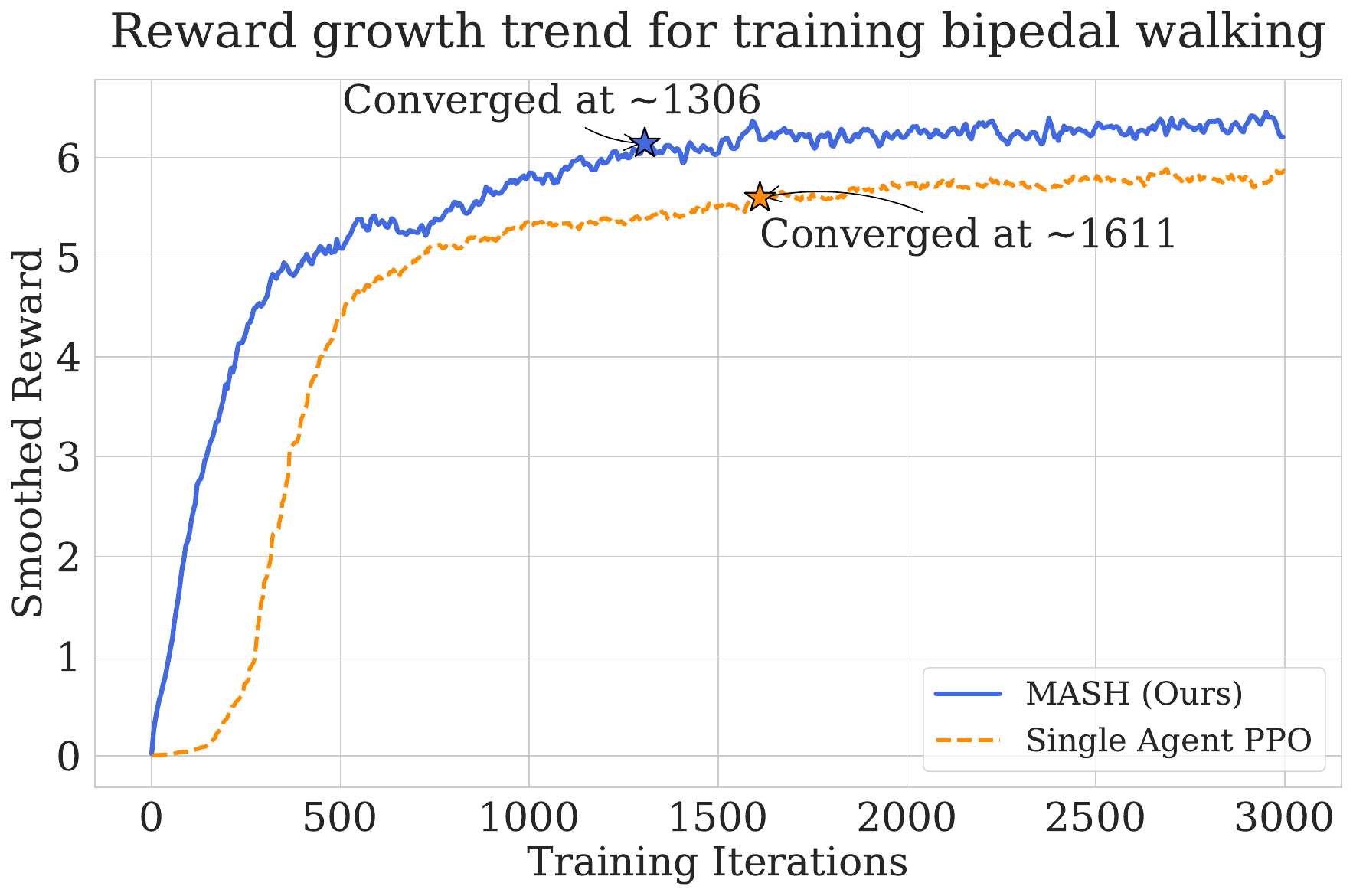} 
    \end{minipage}}
    \subfloat[]{ 
    \begin{minipage}[b]{0.5\textwidth} 
        \centering
        \includegraphics[width=0.9\linewidth]{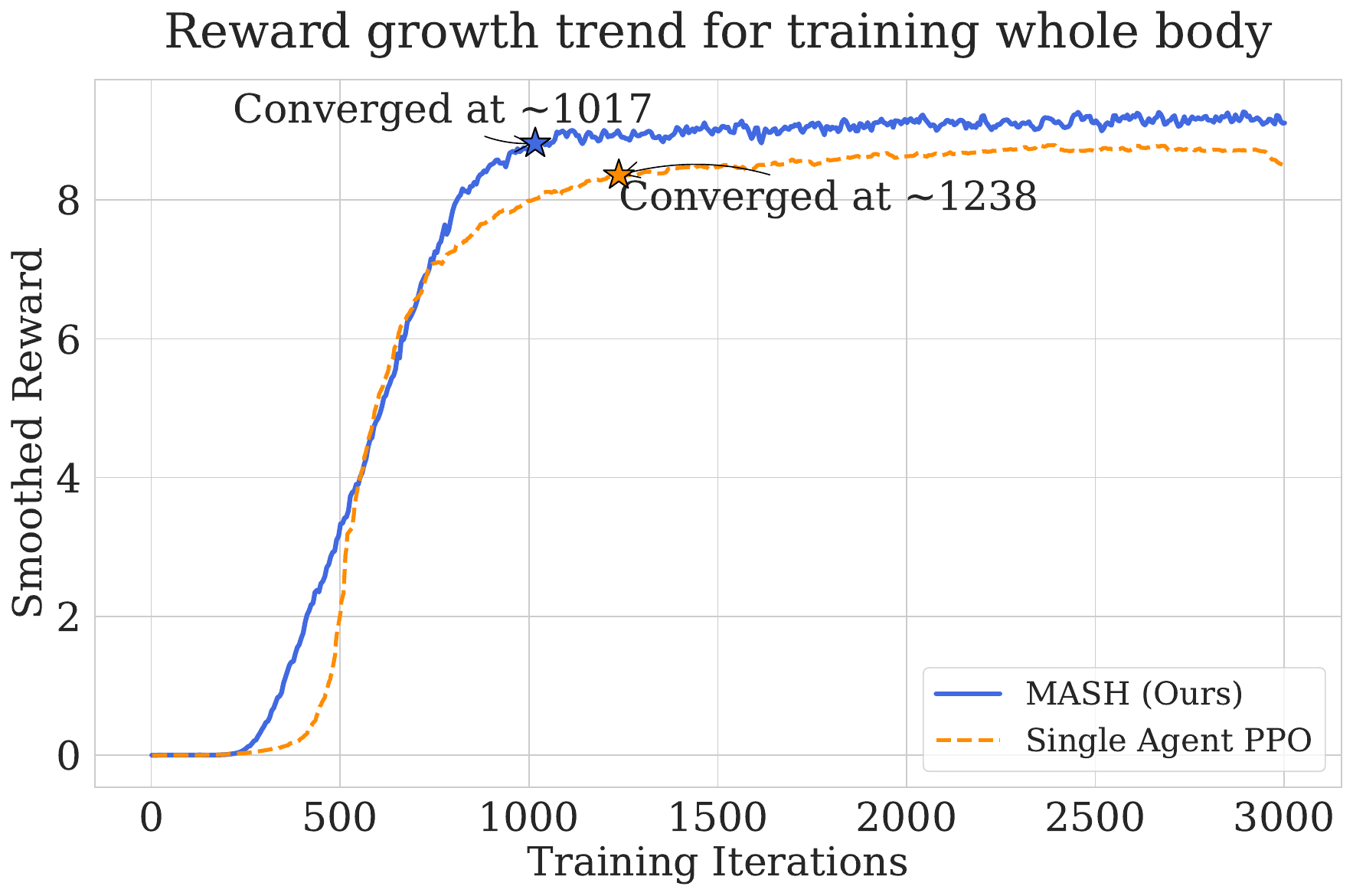} 
    \end{minipage}}
    \caption{Reward growth trends for (a) leg training and (b) whole-body training, comparing MASH with the Single Agent PPO baseline. The smoothed reward is plotted against the number of training iterations.}
    \label{fig: training reward} 
\end{figure*}

\begin{table*}[htbp]
    \centering
    \caption{Evaluation metrics results}
    \label{tab: metrics}
     \renewcommand{\arraystretch}{1.5}
    \begin{tabular}{lcccccccc}
    \toprule
    \multicolumn{1}{c}{\multirow{2}{*}{Method}} & \multicolumn{2}{c}{\(T_{Conv}\)} & \multicolumn{2}{c}{\(S_{action}\)} & \multicolumn{2}{c}{\(S_{torso}\)} & \multicolumn{2}{c}{\(C_{limb}\)} \\
    \cmidrule(lr){2-3} \cmidrule(lr){4-5} \cmidrule(lr){6-7} \cmidrule(lr){8-9}
    \multicolumn{1}{c}{}                        & Bipedal           & Arm-swing        & Bipedal          & Arm-swing          & Bipedal         & Arm-swing         & Bipedal          & Arm-swing          \\ \hline
    MASH (Ours)    & $ \sim$\(\bm{1306}\) (↓) & $ \sim$\(\bm{1017}\) (↓) & \(\bm{0.107}\) (↓) & \(\bm{0.124}\) (↓)  & \(\bm{8.240\times10^{-4}}\) (↓)  & \(\bm{2.720\times10^{-3}}\) (↓)  & \(\bm{0.612}\) (↓) & \(\bm{0.421}\) (↓)                 \\
    Single-agent PPO & $ \sim$\(1661\) & $ \sim$\(1238\) &  \(0.547\) & \(0.546\) & \(3.398\times10^{-3}\) & \(2.802\times10^{-2}\) & \(0.974\) & \(0.951\)                   \\
    \bottomrule
    \end{tabular}
    
\end{table*}

\subsection{Simulation Experiments}
\label{Subsection: Simulation Experiments}
The experimental evaluation compares MASH against the other baselines under identical simulation conditions. Fig. \ref{fig: training reward} illustrates the smoothed reward growth trends during the training process for (a) legs and (b) the whole body, comparing the proposed MASH framework with the Single Agent PPO baseline. In both scenarios, MASH achieves a noticeably faster increase in reward during the early training phase, indicating accelerated learning efficiency. Moreover, MASH converges to a higher asymptotic reward than the baseline, particularly in leg training, where the improvement margin is more pronounced. These results demonstrate the advantage of the multi-agent hierarchical structure in capturing inter-limb coordination patterns and optimizing control performance. After training, we evaluated the deployment performance of all methods. All evaluation metrics are summarized in Table \ref{tab: metrics}. 

As shown in Table \ref{tab: metrics}, we quantitatively evaluate our proposed method, MASH, against a Single-agent PPO baseline across both bipedal and full-body (arm-swing) locomotion tasks. The evaluation metrics—convergence time, action smoothness, torso stability, and limb coordination—demonstrate the clear superiority of our method in nearly all aspects.

\begin{figure*}[htbp]
    \centering
    \centering
    \includegraphics[width=0.8\linewidth]{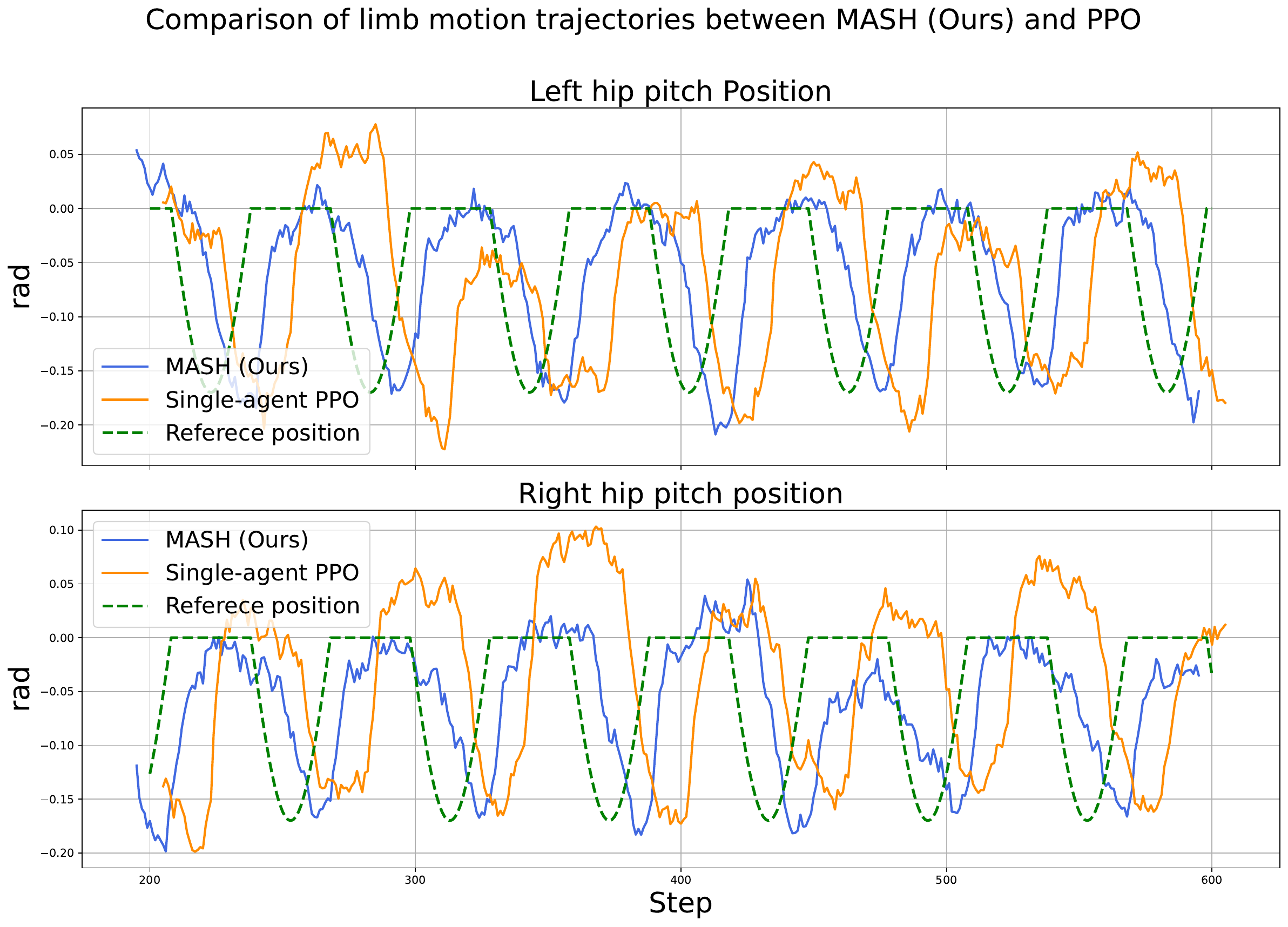} 
    \caption{Comparison of hip pitch joint trajectories. The plots show the tracking performance of our MASH controller versus a Single-agent PPO baseline for the left and right hip joints. Our method (blue) closely follows the reference trajectory (green, dashed), while the baseline (orange) exhibits significant overshoot and phase error.}
    \label{fig: joint pos} 
\end{figure*}

Fig. \ref{fig: joint pos} compares the performance of our MASH controller against a baseline Single-agent PPO on a hip pitch trajectory tracking task. As shown, the PPO controller (orange line) struggles to follow the reference trajectory (green dashed line), exhibiting significant overshooting and phase lag. This results in an unstable and imprecise gait. In contrast, our MASH controller (blue line) demonstrates excellent tracking accuracy, closely matching the reference trajectory in both phase and amplitude. This highlights the superiority of our method in generating stable, smooth, and precise robotic motion.

\subsection{Real-world Experiments}
\label{Subsection: Real-world Experiments}
We deploy MASH on a humanoid robot. To facilitate sim-to-real transfer, we incorporate domain randomization during training, with the parameters listed in Table \ref{tab:domain_randomization}.
\begin{table}[htbp]
\centering
\caption{Domain randomization parameters used in training.}
\label{tab:domain_randomization}
\begin{tabular}{lcc}
\toprule
\textbf{Parameter} & \textbf{Distribution Type} & \textbf{Range / Std. Dev.} \\
\midrule
Friction coefficient       & Uniform  & \([0.1, 1.2]\) \\
Link mass scale factor     & Uniform  & \([0.9, 1.13]\) \\
Center of mass offset (\(m\))  & Uniform  & \([-0.03, 0.03]\) \\
Motor delay (\(ms\))            & Uniform  & \([0.0, 3.0]\) \\
External push force (\(N\))    & Uniform  & \([-20, 20]\) \\
Gravity (\(m/s^2\))        & Uniform  & \([9.78, 9.83]\) \\
Joint damping              & Uniform  & \([0, 0.05]\)   \\
Joint friction             & Uniform  & \([0., 0.05]\)  \\
Joint armature             & Uniform  & \([0.005, 0.015]\) \\
\(K_{p}\) scale factor     & Uniform  & \([0.95, 1.05]\)    \\
\(K_{d}\) scale factor     & Uniform  & \([0.95, 1.05]\)    \\
\bottomrule
\end{tabular}
\end{table}

\begin{figure*}[htbp]
    \centering
    \centering
    \includegraphics[width=0.8\linewidth]{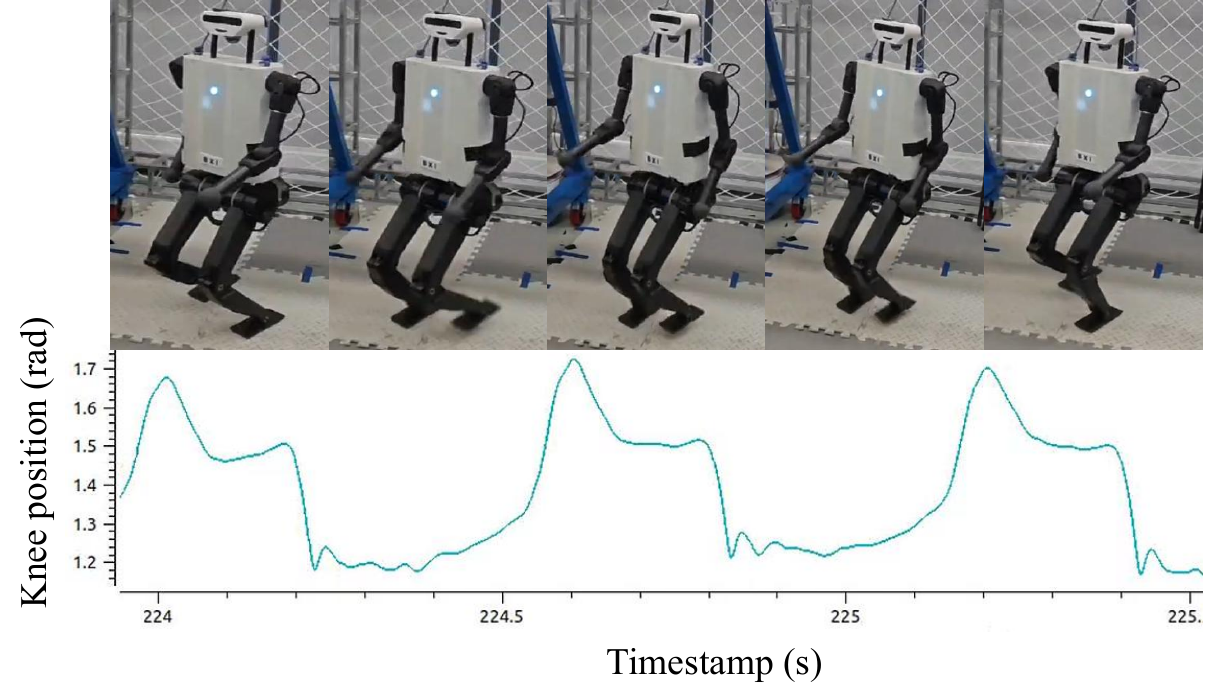} 
    \caption{Stable walking gait on the physical humanoid robot using a policy trained with MASH. The stroboscopic images (top) and corresponding knee joint trajectory (bottom) show a smooth, periodic motion, confirming successful sim-to-real transfer.}
    \label{fig: real} 
\end{figure*}

Fig. \ref{fig: real} demonstrates the real-world hardware validation of a walking gait for a humanoid robot trained using MASH. From a kinematic perspective, the stroboscopic images clearly capture a coherent and fluid gait cycle, showcasing proficient dynamic balance. The corresponding plot of the knee joint angle (in radians) over time (in seconds) quantifies this stable performance. The curve exhibits a high degree of periodicity and smoothness, with the knee angle oscillating regularly within a range of approximately 1.2 to 1.7 radians, and each gait cycle lasting about 0.6 seconds. The smooth, non-abrupt nature of the curve reflects the precision of the motor control and the soundness of the generated policy, effectively avoiding impacts and instability.
This strong correlation between motion and data confirms that MASH can successfully generate a complex motor control policy that is not only effective in simulation but also directly transferable to a real-world physical system. This result indicates that the trained policy possesses excellent robustness, enabling it to handle real-world physical constraints.

\section{Conclusion and Future Work}
\label{Section: Conclusion and Future Work}
In this work, we presented MASH, a novel cooperative-heterogeneous multi-agent reinforcement learning framework designed to enhance the locomotion of a single humanoid robot. By reformulating the control problem and treating each of the robot's limbs as an independent agent within a cooperative system, our approach effectively leverages MARL principles to foster superior inter-limb coordination. Implemented in Isaac Gym, the proposed method demonstrated three key advantages: (1) accelerated training convergence through coordinated limb learning, (2) improving limb movement coordination through multi-agent design, and (3) enhanced robustness via domain randomization. Our experimental results demonstrate that MASH significantly outperforms the conventional single-agent PPO baseline, achieving faster training convergence and a higher asymptotic reward. The learned policies exhibit quantitatively superior performance in terms of action smoothness, torso stability, and limb coordination. Crucially, the efficacy and robustness of our method were validated through the successful sim-to-real transfer of the learned policy to a physical humanoid robot, which executed a stable and smooth walking gait. This study not only introduces a potent and efficient strategy for complex single-robot control but also offers new insights into applying multi-agent learning paradigms to advance humanoid locomotion. For future work, we will apply MASH to robots with other configurations.

\balance

\bibliographystyle{IEEEtran}
\bibliography{main}

\end{document}